\documentclass{article}

\usepackage{PRIMEarxiv}
\usepackage{natbib}[numbers] 
\usepackage[utf8]{inputenc} 
\usepackage[T1]{fontenc}    
\usepackage{hyperref}       
\usepackage{url}            
\usepackage{booktabs}       
\usepackage{amsfonts}       
\usepackage{nicefrac}       
\usepackage{microtype}      
\usepackage{lipsum}
\usepackage{graphicx}
\graphicspath{{media/}}     

\usepackage{hyperref}
\usepackage{url}
\usepackage{amsmath}
\usepackage{amssymb}
\usepackage{booktabs}
\usepackage{graphicx}
\graphicspath{{figures/}}
\usepackage{multirow}
\usepackage{bm}
\usepackage{amsthm}
\theoremstyle{plain}

\theoremstyle{definition}

\theoremstyle{remark}

\usepackage{xcolor}
\definecolor{deltagreen}{rgb}{0.0,0.45,0.0}
\definecolor{deltared}{rgb}{0.6,0.0,0.0}

\definecolor{mixA}{HTML}{0072B2} 
\definecolor{mixB}{HTML}{D55E00} 
\definecolor{mixC}{HTML}{009E73} 
\definecolor{mixD}{HTML}{CC79A7} 
\definecolor{mixE}{HTML}{E69F00} 
\definecolor{mixF}{HTML}{56B4E9} 
\definecolor{mixG}{HTML}{555555} 

\definecolor{distmethod}{HTML}{1D5FA8}
\DeclareMathOperator*{\argmin}{arg\,min}

\title{Quantiles, Not Points:\\ Streaming Quantile Matching for Ordinal-Aware LLM-as-a-Judge Calibration}

\author{
  Ilija Subasic, Andrew Rabinovich, Zhao Chen \\
  Upwork Inc \\
  \texttt{firstname.lastname@\{cloud.\}upwork.com} \\
}

\begin{document}

\maketitle

\begin{abstract}
LLM-as-a-judge scores are often distributionally shifted from human ratings on any given evaluation task, and post-hoc calibration (mapping LLM to human scores in a small labeled anchor set) has become a popular method to correct for this shift. In this paper we derive the calibration from two positions: that LLMaJ scores are ordinal ratings drawn from a distribution rather than isolated points.  We present streaming quantile matching (SQM), a version of a classical method not previously bench marked for LLMaJ calibration, which is built on the two positions by computing the optimal-transport map between the LLM and human score distributions. We develop an evaluation framework and within it we show that unlike point-wise calibrations, SQM respects the ordinality of the human anchor set and achieves the best distributional fidelity and the top overall rank across models and data. We also show that in general, post-hoc calibration narrows the raw error gap between smaller and larger judge models, which opens the possibility of more efficient LLM judges.
\end{abstract}

\section{Introduction}
\label{sec:introduction}

Evaluating large language models (LLMs) outputs
frequently relies on human annotation as a gold standard. Collection of these scores is slow and expensive,
requires recruiting and training raters, and its per-item cost compounds linearly with every new
model, prompt, or use case. Often, this becomes a bottleneck on how fast and safe models can be shipped and monitored.
LLM-as-a-judge (LLMaJ) \citep{li2024survey} has emerged as a scalable substitute: prompting an
LLM to score model outputs. These scores,
however, often disagree with human ratings in systematic ways
\citep{morandi2026debias,zheng2023judging,wu2023styleoversubstance,wang2023notfair}. 

Approaches to address this gap include finetuning
the judge model to match human ratings \citep{zhu2023judgelm},
using reference-free scoring without human input \citep{fu2023gptscore},
or post-hoc calibration of the judge's output \citep{zhao2021calibrate,kadavath2022languagemodels}.
We focus on the last approach, and refer to it simply as calibration, as it retains some of the gold-standard human label quality while not requiring heavy processes like model fine-tuning. Because calibration keeps the weights of the judge model, the same
calibrated judge can be reused as a fixed measurement instrument for
any subsequent analysis.

At its core, we start from the idea that LLMaJ scores are ordinal ratings drawn from a
distribution rather than isolated points, and enforce that any reasonable calibration should respect both the ordinality of the input and the distributions of the human and LLMaJ scores. The clasical
\textbf(quantile matching)~\cite{kolen2014test} method was developed with this in mind, and we adapt it for LLMaJ calibration as streaming quantile matching (SQM). As we lack a robust testing environment for post-hoc calibration, we first build a comparison framework that allows for testing across models, methods, and data budgets (Section~\ref{sec:problem-setup}). Upon testing, we conclude that SQM achieves the top overall rank and wins most often across models, datasets, and processing modes (Section~\ref{subsec:rq1-incremental-vs-full}), and we attribute this to its being both
distribution- and ordinality-aware (Section~\ref{subsec:qual-quantile-wins}). We also observe that calibration narrows the error gap between smaller and larger judge models,
allowing effective use of less resource-intensive judges
(Section~\ref{subsec:rq2-model-gap}), and that the biggest factor in calibration success ends up being the specific dataset being calibrated, with the most promising axes for calibration improvement being clear task definitions and attribute-aware calibration. 
(Section~\ref{subsec:rq3-best-stable}).

\section{Related Work}
\label{sec:related-work}

LLMaJ emerged as a scalable alternative to human evaluation and is widely used, employing pairwise ranking or
rubric absolute scoring \citep{zheng2023judging, liu2023geval}. Despite this success, LLMaJ's sensitivity to prompt design,
model choice, and scoring protocol raises questions about its stability as an
evaluation instrument. Recent surveys structure LLMaJ research along axes such as target
attributes, evaluation protocols, alignment with human judgments, and bias analysis
\citep{gu2024llm, li2024survey, gao2024llm, chang2024survey}. LLMaJ's limitations are well documented: judges systematically disagree with human
raters through strictness offsets and score ranges
\citep{morandi2026debias}, positional and self bias \citep{wang2023notfair,
zheng2023judging}, and sensitivity to surface features such as verbosity and style
\citep{zheng2023judging, wu2023styleoversubstance}.

Three broad strategies have been proposed to
close the resulting gap to human judgment. The first fine-tunes the judge model itself on human preference data, producing dedicated evaluator models such as PandaLM
\citep{wang2024pandalm}, Prometheus \citep{kim2023prometheus}, and JudgeLM \citep{zhu2023judgelm};
this improves alignment but forfeits the frozen-weight property, requiring a new fine-tuning run
whenever the rubric, domain, or backbone model changes. The second replaces a single scalar
judgment with a reference-free scoring function derived from the judge's own likelihoods or
instruction-following behavior \citep{fu2023gptscore}, sidestepping the need for a reference
answer but not the need to correct the judge's own systematic bias. Although allowing for fast iteration,
this approach has limitations in accuracy and explainability outlined by \citet{subasic2026evaluating} and \citet{Bavaresco2024}, and most recently \citet{mukherjee2026geometry}. 
This paper studies a strategy which leaves the judge's weights untouched and instead learns a lightweight 
mapping from raw judge scores to human-aligned scores \citep{zhao2021calibrate,
kadavath2022languagemodels}; unlike fine-tuning, it requires no gradient updates to the judge model, and unlike reference-free scoring, it directly targets the
judge's known bias structure using a small labeled anchor set.

A growing body of work studies  calibration 
taking into account human input and operating under frozen weights regime. \citet{morandi2026debias}~show that small data budgets
lead to good calibration regime. And \citet{rulers2026} report 
that removing the calibration step alone collapses accuracy by 60\%. \citet{calibrate2026dontcurate}
recast calibration as a multi-judge aggregation problem, and \citet{zhu2026finite} asks how much labeled human data a calibrator needs before multi-judge panel is trustworthy. A larger
scale model comparison presented in \citet{emotion2026calibration}  find that isotonic regression best matches the human
label distribution. \citet{verdi2026} and \citet{judgmentdistribution2025} go beyond calibrating the output and 
calibrate a judge's reasoning trace or full output distribution rather
than its point score, and \citet{lager2025} calibrate using cross-layer internal representations
rather than the final output at all.  Concurrently, \citet{steindl2026posthoc} align judges with per-item distributions of annotator labels using trained aligners, whereas we align population-level score distributions with a closed-form, training-free map.
 Each of these targets a different point in the space of
(judge model, calibration method, data budget); none directly compares across all three, which is
the gap this paper addresses.

Stripped of its LLMaJ-specific framing,  calibration under frozen weights is an instance
of the more general problem of matching one distribution to another using only a limited paired
sample: shifting and rescaling a model's output distribution to align with a reference is the
same underlying operation performed by Platt scaling \citep{platt1999} and temperature scaling
\citep{guo2017calibration} for classifier confidence, by \citep{zhao2021calibrate} for language model output probabilities more broadly. What varies across this literature is the
family of transport map allowed (affine, monotone, or fully non-parametric)~(\citep{ma2026distributional} compares for reward functions), the loss used to
measure distributional discrepancy (mean error, KL divergence, Wasserstein distance)~\citep{niculescu2005predicting}, and how much
paired data the map is fit on~\citep{lee2026instanceoptimal}; the methods compared in this paper span this same design space, applied uniformly to the single problem of aligning a frozen LLM judge's scores
with human ratings. Unlike this line of work, which treats calibration as fitting a mapping between two sets of point scores, 
we start from the view that LLM-judge scores are ordinal ratings inspired by~\citep{kuleshov2018accurate} drawn from a distribution and design methods around respecting both. 
We adapt classical quantile (equipercentile) matching~\citep{kolen2014test}, estimated online with the {$P^2$} estimator~\citep{jain1985p2}, to LLM-judge calibration (streaming quantile matching, SQM).


\section{Problem Setup}
\label{sec:problem-setup}

We first fix the terminology used throughout the rest of the paper, then formalize the
frozen-weight calibration problem introduced in Section~\ref{sec:introduction} and describe the
held-out evaluation protocol and metrics used to compare methods.

\paragraph{Terminology.} A \textbf{model} is the underlying frozen LLM judge backbone that produces a raw score for a given input; its
weights are never updated anywhere in this paper. A \textbf{prompt} is the fixed instruction and
scoring rubric the model is given at inference time to produce that raw score (e.g.\ G-Eval \citep{liu2023geval} per-attribute rubrics used for SummEval); changing
the prompt changes what the model is being asked to do, but never its weights. A \textbf{method}
is a calibration function: a mapping from a model's raw score(s) to a human-aligned score, fit on
a small set of labeled anchors and applied afterward with no further access to the model. A
\textbf{mode} is the regime under which a method is fit to its anchor set: \emph{streaming} fits
and re-fits on one labeled example at a time; \emph{batch} fits and re-fits on fixed-size
batches of $B$ examples; and \emph{full-batch} fits once on the entire available calibration pool in a
single shot, with no incremental updates. Streaming and batch modes are re-evaluated at
every checkpoint as more data arrives, while full-batch is fit and evaluated exactly once.

\paragraph{Frozen-weight calibration.} For a fixed model $m$ and prompt
$p$, let $j = m_p(x) \in \{n \in \mathbb{Z} \mid a \le n \le b\}$ denote the raw judge score assigned to input $x$,
and let $h(x) \in \{n \in \mathbb{Z} \mid a \le n \le b\}$ denote the corresponding human score.
A method is a function $g_\phi: \{n \in \mathbb{Z} \mid a \le n \le b\} \to \mathbb{Z}$,
parameterized by $\phi$, that is fit on a labeled anchor set $\mathcal{A} = \{(j_i, h_i)\}_{i=1}^n$
so that $g_\phi(j)$ approximates $h$ as closely as possible; at inference time $g_\phi$ is applied
to $m_p(x)$ for new, unlabeled $x$, with $m_p$ itself never updated. Following the
distribution-matching view, if $P_J$ is the
distribution of raw scores $j$ induced by $m_p$ over a population of items and $P_H$ is the
corresponding distribution of human scores $h$ over the same items, a method's goal is to find
$g_\phi$ such that the pushforward distribution $g_\phi(P_J)$ is as close as possible to $P_H$,
without access to $P_H$ beyond the $n$ paired anchors in $\mathcal{A}$.

\paragraph{Held-out evaluation protocol.} For every (dataset, attribute, model) group, we reserve
a fixed held-out fraction of the data (e.g.\ $0.3$) that is never used
to fit any method; the remaining pool is what full-batch mode fits on in one shot, and what
streaming/batch modes replay incrementally. Every method and mode in a group is
therefore scored against the exact same held-out population, at whatever amount of calibration
data it has seen so far: full-batch mode is evaluated once, after its single fit on the entire
non-held-out pool, while streaming/batch modes are re-scored against the held-out
set at every checkpoint as they see more of the (non-held-out) data. This makes comparisons across
modes and data budgets apples-to-apples: a streaming method that has seen 20\% of the calibration
pool and a full-batch method fit on exactly that same 20\% are both scored on the identical
held-out items, so differences in reported error reflect the fitting method and data budget, not
which items happened to be evaluated.

\paragraph{Metrics.} Our primary metric is \textbf{held-out mean absolute error}: the mean of $|g_\phi(j_i) - h_i|$
over the fixed held-out set, computed
using whatever state a method has fit so far. This is distinct from, and preferred over, the
naive \textbf{cumulative MAE}, the running average of a streaming/batch method's error on
each new example \emph{as it is seen}. Cumulative MAE is
biased upward for as long as it includes a method's early, poorly-fit cold-start period, no matter
how far into the stream it is measured. Held-out MAE has no such bias:
it is always an independent, current-state evaluation of the method's present fit against data it
has never been fit on, so it is comparable across methods, modes, and checkpoints without any
correction. We additionally care about matching the \emph{shape} of the score distribution rather than
only its mean absolute deviation from human scores. For instance, whether a method reproduces
human raters' full spread of scores rather than only their average. For this we report the
\textbf{Earth Mover's Distance (EMD)} \citep{rubner2000earthmover}, also known as the
1-Wasserstein distance, between the calibrated score distribution and the human score distribution
on the held-out set, $W_1(\hat{P}_H, \hat{P}_{g_\phi(J)})$. We prefer EMD over
Kullback--Leibler (KL) divergence \citep{kullback1951information} because
it is an ordinal-aware metric: it weights each unit of probability mass by how far it must be moved
along the score scale to match the target distribution, so a method that misplaces mass onto scores
close to the correct one is penalized less than a method that misplaces the same mass onto scores far
away. KL divergence, by contrast, treats scores as unordered categorical outcomes and penalizes any
mismatch between the two distributions equally regardless of how close the scores are on the
underlying Likert scale. A method can achieve a low held-out MAE while still mismatching the human
distribution's shape (e.g.\ by regressing every prediction toward the population mean), and EMD is
what distinguishes that failure mode from genuine distributional alignment, while remaining
sensitive to the ordinal structure that MAE alone discards at the level of aggregate shape. Finally, for
clarity and understanding we report the \textbf{exact-match rate} on the held-out set,
the fraction of items for which the calibrated and human score are identical.

\subsection{Datasets}
\label{subsec:datasets}

We evaluate on five datasets spanning judge rubrics of differing granularity.
\textbf{BiGGen Bench} \citep{kim2024biggenbench} defines nine capabilities with
instance-specific rubrics; in our pipeline these are scored under a single overall
attribute using the benchmark's released absolute-scoring prompt template. \textbf{FLASK}
\citep{ye2024flask} covers prompts annotated against named skill rubrics (e.g.\ logical
robustness, factuality, commonsense understanding), using its released skill-specific judge
prompts, collapsed to a single \texttt{quality} attribute in our pipeline. \textbf{HelpSteer2}
\citep{wang2024helpsteer2} provides 5 attributes (helpfulness, correctness, coherence,
complexity, verbosity). HelpSteer2 ships no released LLM judge prompt; we reconstruct one from
its annotator guidelines, and flag this dataset's judge-prompt source as reconstructed rather
than released throughout. \textbf{SummEval} \citep{fabbri2021summeval}, scored via the
\textbf{G-Eval} prompts \citep{liu2023geval}, provides 4 attributes (coherence, consistency,
fluency, relevance) on (model, summary) pairs from CNN/DailyMail articles.
\textbf{UPHELD} \citep{subasic2026evaluating} is our fifth dataset, added to broaden coverage  to multi-turn, free-form dialogue: human
raters score a candidate assistant response given its preceding conversation turns. All five
datasets use a five point integer Likert scale, making them directly
comparable for calibration.

\begin{table}[t]
\footnotesize
\centering
\caption{Dataset scale used for calibration. \emph{Initial data points} is the ground-truth
count of distinct underlying items
before any candidate response or judge is involved. \emph{Raw judgements} is the total number of
individual judge scores our pipeline produces from a dataset.}
\label{tab:dataset-stats}
\begin{tabular}{lrr}
\toprule
Dataset & Initial data points & Raw judgements \\
\midrule
BiGGen Bench & 695    & 9{,}916  \\
FLASK        & 168    & 6{,}429  \\
HelpSteer2   & 519    & 17{,}304 \\
SummEval     & 100    & 24{,}149 \\
UPHELD       & 408    & 4{,}823  \\
\midrule
Total        & 1{,}890 & 62{,}621 \\
\bottomrule
\end{tabular}
\end{table}

Table~\ref{tab:dataset-stats} reports, for each dataset, the number of \emph{initial data
points}, the ground-truth count of distinct items (instructions, source
articles/conversations, or prompts) before any candidate response or judge is involved, and
the total number of \emph{raw judgements} our pipeline produces from them. The gap between the
two columns comes from two multiplicative factors: (i) each initial item is paired
with more than one candidate response to be scored (e.g.\ SummEval's 100 articles each have 16 candidate summaries, giving
$1{,}600$ pairs), and (ii) every one of those (item, response) pairs is then
scored by four judge models (Section~\ref{subsec:models}), and, for
HelpSteer2 and SummEval, once per judged dimension on top of that, since each of the dimensions is an
independently fit calibration target. A fraction ($\leq2\%$) of judge calls fail to
return a parseable score and are dropped rather than imputed.

\subsection{Models}
\label{subsec:models}

We calibrate four judge backbones, spanning both closed- and open-weight, and both
larger and smaller parameter counts: \textbf{GPT-5} \citep{openai2025gpt5}, \textbf{Kimi K3}
\citep{moonshotai2026kimik3}, a 2.8T-parameter open-weight MoE model from Moonshot
AI, \textbf{GPT-OSS-20b} \citep{openai2025gptoss}, OpenAI's smaller open-weight model, and
\textbf{Qwen3-30B} (specifically a3b-instruct-2507) \citep{qwen2025qwen3}. All
models are queried through a fixed inference gateway with weights entirely frozen throughout
every experiment (we have no control over GPT-5's weights, but our experiments were run consecutively and averaged);
GPT-OSS-20b and Qwen3-30B serve as the smaller judges compared against GPT-5 and Kimi K3 as the larger, frontier judges in the model-gap analysis of
Section~\ref{sec:results}.

\section{Calibration Methods}
\label{sec:methods}

Table~\ref{tab:rq1-heldout-medians} reports each method we compare and their fitting mode (\emph{Streaming} (fit and re-fit on one anchor at a time); \emph{Batch} (fit and re-fit on fixed-size batches); 
\emph{Full-batch} (fit once on the entire non-held-out pool)).  Formal definitions of every method not given a formula below are in
Appendix~\ref{app:formal-definitions}.

\subsection{SQM for LLMaJ Calibration}
\label{subsec:methods-distributional}

\textbf{Streaming quantile matching} incrementally aligns the judge's empirical score quantiles to
the human score quantiles, using the online $P^2$ quantile estimator \citep{jain1985p2} so that
neither distribution needs to be stored in full (on a K-point scale the empirical CDFs are maintained exactly with K running counts). With $\hat{F}_J, \hat{F}_H$ the empirical CDFs of
the judge and human scores among anchors seen so far,
\[
g_\phi(j) = \hat{F}_H^{-1}\!\left(\hat{F}_J(j)\right).
\]
The point-wise methods fit $g_\phi$
against the paired anchors $(j_i, h_i)$ directly. 

\textbf{SQM over predictive score distributions} extends SQM to judges that expose a full per-example predictive distribution over ratings rather than a single
collapsed score. This can be done either as
explicit categorical $(v, w)$ over the rating enum (e.g.\ the softmax over the score token's logits)
or an unweighted sample from a rollout of repeated stochastic completions. Each example's entire distribution, not just its mean or argmax, is pooled as
weighted mass into the running judge distribution, so $\hat{F}_J$ reflects the judge's predictive
uncertainty at every score level and is a lower-variance estimate of the judge's marginal than an
equal-count sample of point scores would give. A query's own predictive distribution is then
calibrated pointwise, support point by support point,
\[
g_\phi(v) = \hat{F}_H^{-1}\!\left(\hat{F}_J(v)\right) \quad \text{for each } v \in \operatorname{supp}(j),
\]
with the original weights $w$ preserved, so the returned calibrated distribution keeps the query's
original shape (spread, skew, multi-modality) instead of collapsing it to a point before calibration.
Taking the weighted mean of this calibrated distribution recovers a single point estimate, and a
bare scalar judge score degenerates to a one-point distribution of weight 1 -- so this method is a strict
generalization of, and a drop-in replacement for, plain streaming quantile matching.

\subsection{Other CDF Methods}

\textbf{Mini-batch sliced Wasserstein optimal transport} aligns the judge and human score
distributions batch-by-batch \citep{bonneel2015sliced}. For a mini-batch of size $B$, sorting both
sides to order statistics $j_{(1)} \le \dots \le j_{(B)}$ and $h_{(1)} \le \dots \le h_{(B)}$ gives
the closed-form 1-D optimal-transport plan,
\[
\{(j_{(i)}, h_{(i)})\}_{i=1}^B = \argmin_{\sigma \in S_B} \sum_{i=1}^B \left(j_{(i)} - h_{\sigma(i)}\right)^2;
\]
matched pairs from every mini-batch accumulate into a growing reference set $\mathcal{R}$, and
$g_\phi(j)$ is a piecewise-linear interpolation of $\mathcal{R}$ at $j$.

\textbf{Cross-entropy matching} restricts $g_\phi$ to an affine map $g_\phi(j) = aj + b$ and fits
$(a,b)$ against the two marginals' histograms $p(a,b)$ and $q$ over the score range, rather than
against paired anchors,
\[
(a, b) = \argmin_{a,b} \operatorname{KL}\big(p(a,b) \,\|\, q\big).
\]

SQM's nonparametric quantile-to-quantile map is the closed-form solution to
the 1-D optimal-transport problem \citep{peyre2019computational} under the $L_1$/EMD cost -- exactly the discrepancy Earth Mover's
distance measures (Section~\ref{sec:problem-setup}) -- while cross-entropy matching restricts the
map to an affine rescaling under KL divergence, a non-ordinal-aware objective. Both restrict
themselves to the two marginal distributions, but only one of them is optimizing the same
discrepancy our evaluation rewards; Section~\ref{subsec:qual-quantile-wins} returns to this
contrast and why it decides which method wins.

\section{Results}
\label{sec:results}

All our experiments are run on a 30\% held-out dataset, running each setup three times to account for
randomness in LLM-judge scores and potential parsing errors. We structure the experiments through a set
of questions we pose and answers we distill from the experimental results.

\subsection{Which Methods Reduce Error Relative to the Raw Judge?}
\label{subsec:rq1-does-calibration-help}

\begin{figure}[t]
\centering
\includegraphics[width=\textwidth]{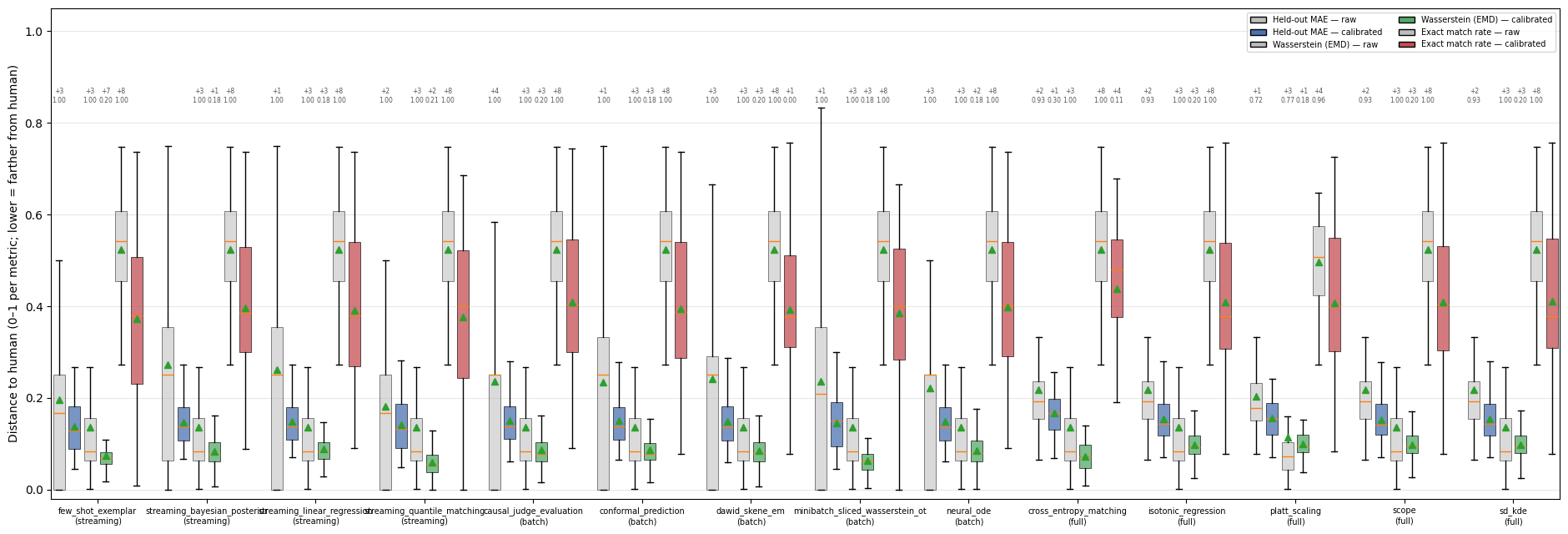}
\caption{Each metric min--max
normalized to $[0,1]$ and exact-match rate expressed as a mismatch rate, so that lower is always
closer to human on every metric.}
\label{fig:rq1-appendix-normalized}
\end{figure}

Prior work has shown the effectiveness of calibration \citep{morandi2026debias, rulers2026,
calibrate2026dontcurate, emotion2026calibration}. We first confirm this holds in our setup, and Table~\ref{tab:heldout-medians-subset} a
reports the improvement over the raw scores (Table~\ref{tab:rq1-heldout-medians} holds full results), pooled across datasets, models, and prompts. Figure~\ref{fig:rq1-appendix-normalized} reports the same scores normalized to a 0-1 range. Using a paired Wilcoxon
signed-rank test (Holm-corrected within each method's family of comparisons), we find a
statistically significant improvement over the raw judge in 14 of the 17
methods, and the best held-out checkpoint of every method beats the raw judge on at least one
metric. Calibration left rank correlation with human judgments unchanged: the median Spearman $\rho = 0.494$ and Kendall $\tau_b = 0.450$ are the same before and after calibration, pooled across methods and datasets. For monotone calibrators (SQM, isotonic regression, Platt scaling, affine maps) this holds by construction, since a non-decreasing map cannot reorder items. That it also holds empirically for the non-monotone methods indicates that post-hoc calibration corrects the judge's scale and distribution rather than its ranking of items. This is analogous to temperature scaling, which reduces calibration error while leaving accuracy unchanged~\citep{guo2017calibration}.

\begin{table}[t]
  \centering
  \small
  \caption{Held-out \textbf{Median} performance pooled across datasets, models, and prompts for SQM, CDF, and top baseline methods. \textbf{Bold} indicates best performance; \underline{underline} marks second best; ($\Delta$\,\%) denotes relative improvement over raw scores; \emph{Rank (W/L)} denotes overall pairwise majority-vote rank and wins/losses across all 17 evaluated methods.}
  \label{tab:heldout-medians-subset}
  \begin{tabular}{llrrrl}
  \toprule
  Method (Reference) & Mode & MAE $\downarrow$ & Wasserstein $\downarrow$ & Exact Match $\uparrow$ & Rank (W/L) \\
  \midrule
  \multicolumn{6}{l}{\textbf{SQM methods} (Our Proposal)} \\
  \midrule
  Streaming quantile matching & Streaming & \underline{0.53 (+31.37)} & \underline{0.26 (+27.87)} & 0.55 (+30.86) & \textbf{1 (7W/0L)} \\
  SQM over predictive score distributions & Streaming & 0.53 (+30.81) & \textbf{0.24 (+33.80)} & 0.55 (+30.73) & 6 (2W/0L) \\
  \midrule
  \multicolumn{6}{l}{\textbf{CDF methods}} \\
  \midrule
  Sliced Wasserstein OT \citep{bonneel2015sliced} & Batch & 0.60 (+22.45) & 0.29 (+20.52) & 0.55 (+30.82) & 5 (5W/0L) \\
  Cross-entropy matching & Full-batch & 0.67 (+13.36) & 0.32 (+11.47) & 0.47 (+13.49) & 17 (0W/3L) \\
  Contrastive rollout calib. \citep{rimsky2023caa} & Streaming & 0.65 (+16.07) & 0.34 (+7.71) & 0.48 (+15.48) & 15 (0W/1L) \\
  \midrule
  \multicolumn{6}{l}{\textbf{Best among other methods}} \\
  \midrule
  Dawid--Skene EM \citep{dawidskene1979} & Batch & 0.54 (+29.79) & 0.35 (+4.00) & \textbf{0.57 (+35.41)} & \underline{2 (1W/0L)} \\
  Few-shot exemplar \citep{zhao2021calibrate} & Streaming & \textbf{0.52 (+32.86)} & 0.31 (+15.27) & \underline{0.56 (+35.16)} & 3 (5W/1L) \\
  \bottomrule
  \end{tabular}
\end{table}

\subsection{Which methods reach their best score faster?}
\label{subsec:rq1-incremental-vs-full}

\begin{figure}[htb]

\centering
\includegraphics[width=15cm, height=9cm, keepaspectratio=true]{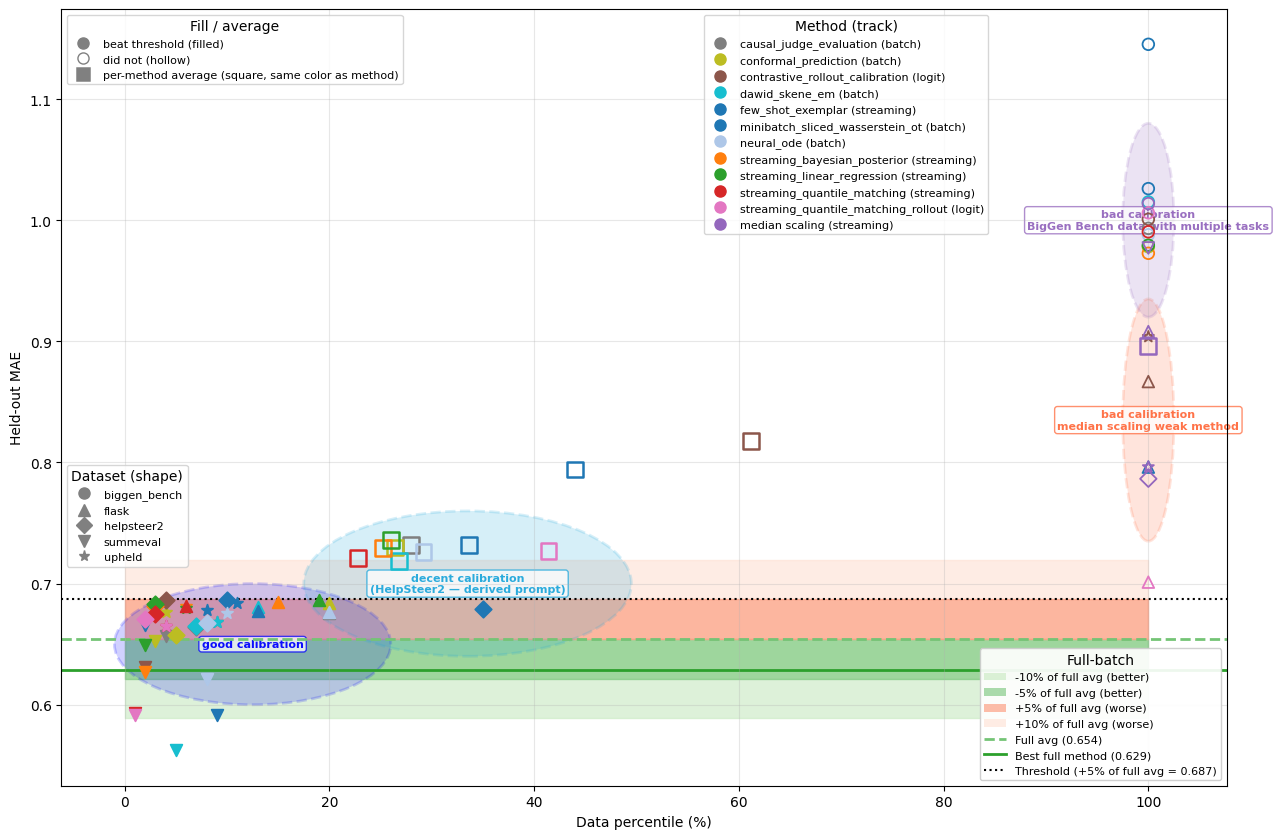}

\caption{For each (method, dataset) pair: held-out MAE at the first point in the stream where it falls within +5\% of the full-batch reference (filled = reached; hollow = never reached, plotted at 100\%; squares = per-method averages). Shaded bands mark $\pm$5\% (darker) and $\pm10$\% (lighter) around the full-batch average.}
\label{fig:rq1-earlystop-mae}
\end{figure}

We ask whether the SQM incremental
distribution-matching methods reach a good calibration faster than fitting a method once on the
full non-held-out pool. Figure~\ref{fig:rq1-earlystop-mae} shows an early-stop view, the first percentile at which each method's per-dataset curve enters a $+5\%$ band around the full-batch average (0.65; best full-batch method, SCOPE, at 0.63). SQM is among the methods which cross the threshold the earliest, entering the band on SummEval at roughly the first percentile of the pool and
on UPHELD by roughly the ninth. Mini-batch sliced Wasserstein OT does not visibly share this
advantage: because it only updates once per fixed-size batch
rather than refining a single running estimate after every anchor, its data-efficiency is
structurally coarse. 

In terms of actual run time, SQM falls in the top-3 methods, with only Streaming linear regression and Isotonic regression beating it at 0.7 and 0.8$\times$ the time needed for the SQM to finish. The slowest method was SD-KDE which on average took $\approx9\times$ more than SQM. Note that all the times for a single calibrations are recorded in less than 10ms (run on a MacBook Pro M5, 48GB RAM laptop). The predictive-distribution variant of SQM is slower, taking about $50\times$ as long as single-score SQM on average, because it must parse token probabilities.

\subsubsection {How much data is needed?}
\begin{figure}[htb]
\centering
\includegraphics[width=0.7\textwidth]{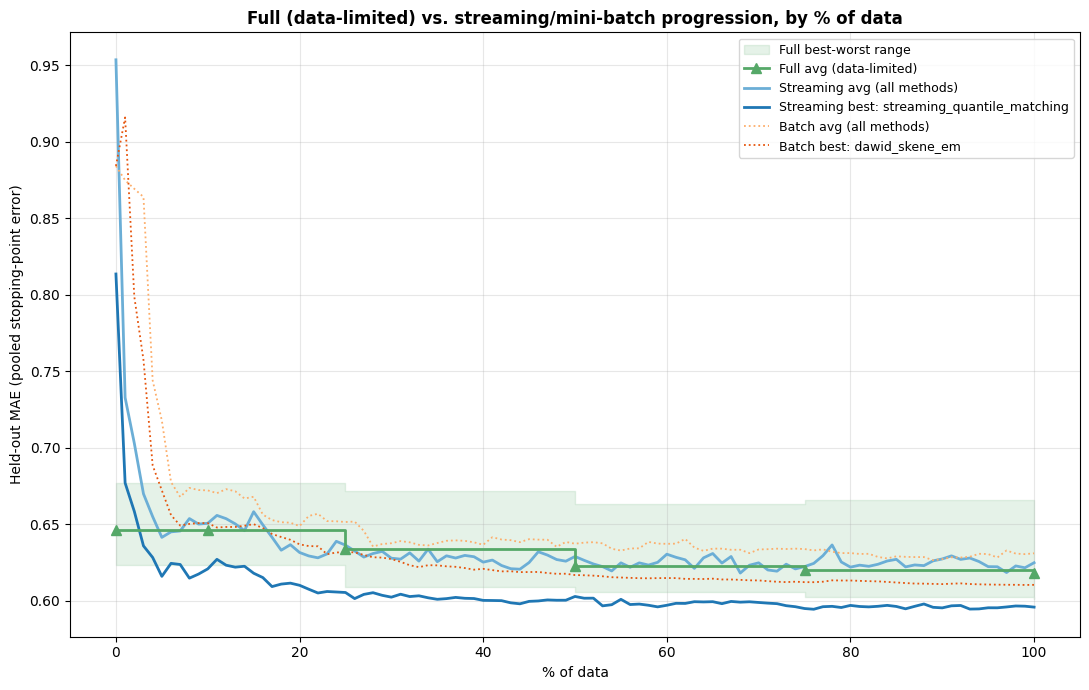}
\caption{Full-batch method fit on limited amounts of training data (10/25/50/75/100\%, green
step function with best--worst range shaded) versus streaming's and batch's own
pooled stopping-point curves across every percentile of data (best method and across-method
average for each track). Held-out MAE, pooled across all four judge models.}
\label{fig:rq1-full-limited}
\end{figure}

A related question is whether methods need all of its data: fitting it on
just the first 10/25/50/75\% of the
training pool shows how much
held-out MAE improves with a tenfold increase in data. Figure~\ref{fig:rq1-full-limited}  reports this comparison; streaming quantile matching's own
pooled trajectory (average $0.61$) beats every one of these data-limited full-batch checkpoints,
including the checkpoint fit on $100\%$ of the data, using far less labeled data to get there.

\subsubsection{Automatic Stopping Criteria}
\label{app:rq1-3-auto-stop}

Section~\ref{subsec:rq1-incremental-vs-full} reports only the headline conclusion that fixed-horizon
stopping outperforms adaptive stopping; this section gives the full comparison behind that
statement. We compare five loss-aware adaptive stopping rules -- an EMA-slope detector
(\texttt{ema\_slope}), a Page--Hinkley change-point test (\texttt{page\_hinkley}), the ADWIN
sliding-window drift detector (\texttt{adwin}), a champion-challenger holdout comparison
(\texttt{champion\_challenger}), and a reservoir-validation check (\texttt{reservoir\_validation})
-- against two fixed-horizon baselines that simply stop after exactly 100 or 200 examples
(\texttt{stop\_at\_100}, \texttt{stop\_at\_200}), for every streaming and batch method, dataset, and
judge model.

\begin{figure}[htb]
\centering
\includegraphics[width=\textwidth]{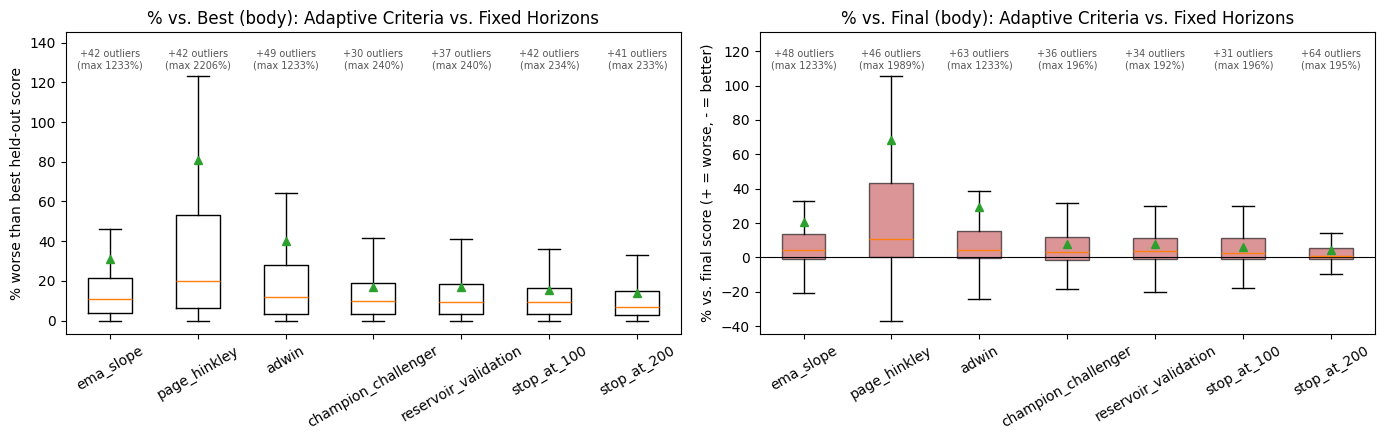}
\caption{Automatic stopping rules compared on two regret measures, pooled across all streaming and
batch methods, datasets, and judge models. \textbf{Left:} \% worse than that run's own best-ever
held-out score (lower is better). \textbf{Right:} \% vs.\ the run's eventual final score after
consuming the full pool (negative means the stopping point already beat the eventual final score).
Orange line: median; green triangle: mean; outlier counts and the most extreme outlier are
annotated above each box. The two fixed-horizon rules (\texttt{stop\_at\_100},
\texttt{stop\_at\_200}) have the lowest median and mean regret of all seven rules on both
measures.}
\label{fig:rq1-stopping}
\end{figure}

Figure~\ref{fig:rq1-stopping} scores each rule two ways: how much worse its stopping point's
held-out score is than that same run's own best-ever checkpoint (left), and how much worse it is
than the score the run eventually reaches if left to consume the full pool (right; negative is
better than the eventual final score). On both measures, the two fixed horizons dominate: median
regret is smallest for \texttt{stop\_at\_200} (roughly 5--6\% worse than the best checkpoint,
and effectively on par with the eventual final score) and \texttt{stop\_at\_100} (roughly
9\% and 2\% respectively), while every adaptive rule has a higher median regret and a
much heavier tail of outliers. \texttt{page\_hinkley} is the least reliable of all seven rules,
with a mean regret an order of magnitude above its own median (driven by runs where it fires far
too early or never fires at all) and outliers exceeding 1200--2200\% worse than the best checkpoint.
\texttt{champion\_challenger} and \texttt{reservoir\_validation} are the best-behaved of the five
adaptive rules but still do not beat either fixed horizon on median or mean regret. In short,
knowing the loss curve's shape in advance (a fixed horizon) is a more reliable stopping strategy in
our setting than trying to detect convergence from the stream itself.

\subsection{Which methods are more stable across datasets?}
\label{subsec:rq3-best-stable}

Section~\ref{subsec:rq1-does-calibration-help}'s pooled numbers can hide a method that wins big on
one metric or dataset and especially badly on another. Figure~\ref{fig:rq1-earlystop-mae} makes
this visible directly: plotting every (method, dataset) pair's early, and stopping percentile against
its held-out MAE does not separate by method. Instead they form into four clusters that are mostly
explained by the \emph{dataset} a point belongs to.

The majority of (method, dataset) pairs FLASK, HelpSteer2, SummEval, and UPHELD, across
essentially every method we test fall into one
\emph{good calibration} cluster, reaching the full-batch reference's $+5\%$ band within the first
$0$--$25\%$ of the labeled pool. A second cluster, \emph{decent calibration (HelpSteer2 -- derived
prompt)}, is not made of individual points but of several methods' per-dataset \emph{averages}
sitting right at that threshold; this is a property of HelpSteer2, which ships no released
LLM-judge prompt and is scored with one we reconstruct (Section~\ref{subsec:datasets}), not a
weakness of the methods themselves, whose individual points on the other three datasets sit
well inside \emph{good calibration}. A third cluster, \emph{bad calibration -- BigGen Bench data
on multiple tasks}, contains essentially every method's BigGen Bench point, SQM's included: none of them ever reaches the threshold 
on this dataset, suggesting this is a property of the
dataset, not of any one method's fit. Only the fourth cluster is a genuine method-level exception to this pattern, and it cuts against
the same two methods in two different ways. Median-scaling's points on
FLASK, HelpSteer2, and UPHELD land in a cluster of their own, \emph{bad calibration -- median
scaling weak method}, well outside \emph{good calibration}: it is the one method that fails to
calibrate well even on the datasets every other method finds easy.

SQM's stability, read this way, is not that it stands apart from
every other method. What sets it
apart is that it never becomes an outlier in the bad direction: its one weak dataset, BigGen Bench,
fails exactly as much as the dataset forces every method to fail, never more, unlike few-shot
exemplar's catastrophic BigGen Bench point or median scaling's separate cluster of
failures on all datasets.

\subsection{Which models are calibrated better?}
\label{subsec:rq2-model-gap}

\begin{figure}
    \centering
    \includegraphics[width=0.95\linewidth]{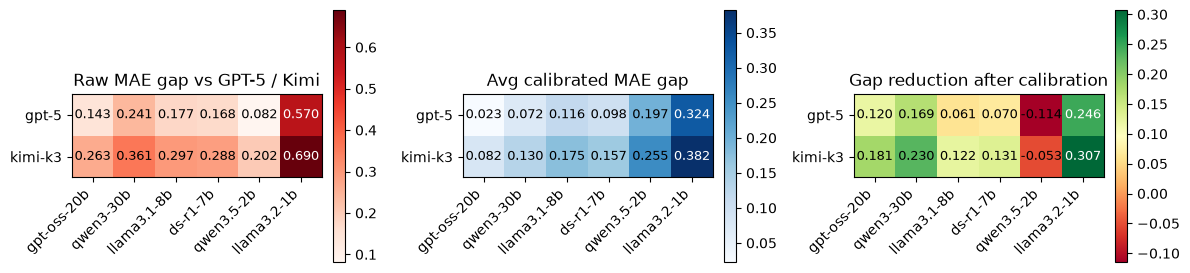}
    \caption{Heat map of calibration gap between frontier and medium, and smaller model.}
    \label{fig:heatmap}
\end{figure}

Prior work~\citep{lail2026costeffective} showed that ensembles of ``smaller''
judges can match the performance of ``larger'' judges; we ask whether calibration
can achieve the same. We compare eight judges: two ``large'' (GPT-5, Kimi-K3),
two ``medium'' (GPT-OSS-20b, Qwen3-30b), and four ``small'' (Llama3.1-8b,
DeepSeek-R1-7b (ds-r1), Qwen3.5-2b, Llama3.2-1b). Figure~\ref{fig:heatmap} shows,
for each medium and small judge, its MAE gap to each large judge before
calibration (left), after calibration averaged over all methods (middle), and the
resulting reduction (right). The reduction is far from uniform.

Before calibration, gaps range from 0.082 to 0.570 relative to GPT-5 and from
0.202 to 0.690 relative to Kimi-K3. Llama3.2-1b is by far the furthest from both
large judges (0.570 and 0.690), whereas Qwen3.5-2b is the closest (0.082 and
0.202), so the raw gap does not simply follow model size.

On average, calibration brings the medium judges closest to the large judges. After
averaged calibration, GPT-OSS-20b and Qwen3-30b are within 0.023 and 0.072 of GPT-5
and within 0.082 and 0.130 of Kimi-K3. This corresponds to closing 83.9\% and
70.1\% of their raw gap to GPT-5 and 68.8\% and 63.7\% of their raw gap to Kimi-K3.
The small judges that improve close a smaller fraction, 34.5--45.5\% of their raw
gaps (Llama3.1-8b: 34.5\%/41.1\%; ds-r1: 41.7\%/45.5\%; Llama3.2-1b: 43.2\%/44.5\%,
for GPT-5/Kimi-K3). Llama3.2-1b's absolute reductions (0.246 and 0.307) are the
largest in the figure, but it remains the furthest from both large judges after
calibration (0.324 and 0.382). The medium judges also do not reach parity with every
large judge: Qwen3-30b remains 0.130 from Kimi-K3.

This advantage is not monotonic in model size. Calibration \emph{widens} the gap for
Qwen3.5-2b, the closest small judge before calibration, from 0.082 to 0.197 relative
to GPT-5 (a roughly 140\% increase) and from 0.202 to 0.255 relative to Kimi-K3
(a 26.2\% increase). We note that Qwen3.5-2b also had a markedly higher error rate
when processing its outputs (up to 20\%, versus under 3\% for all other judges), so
this result may partly reflect output-processing failures rather than calibration
alone. In addition, the closeness of GPT-OSS-20b to GPT-5 (0.023, versus 0.082 to
Kimi-K3) may reflect a shared model family rather than size alone, and with only two
medium judges we cannot separate these effects.

 \subsection{Why does streaming quantile matching win?}
\label{subsec:qual-quantile-wins}

SQM ranks first overall and its two variants achieve the lowest median Wasserstein distances of all methods (0.24 for the predictive-distribution variant and 0.26 for single-score SQM, vs. 0.29–0.43 for all others; Table~\ref{tab:rq1-heldout-medians}). Direct-fit streaming baselines constrain the correction $g_\phi$ to fixed parametric families (e.g., affine maps, quadratic features, or local exemplars). In contrast, SQM's nonparametric quantile-to-quantile map directly aligns empirical CDFs---the exact closed-form optimal-transport solution under the $L_1$/EMD objective used to measure distributional success. This distributional fit also carries over to pointwise metrics: SQM is within 0.01 of the best median MAE and 0.02 of the best median exact-match rate. We hypothesize that matching the full score distribution, rather than fitting a fixed parametric form, makes SQM less sensitive to rubric differences across tasks.

Extending this nonparametric mapping to full example score distributions rather than a single collapsed value (Section~\ref{subsec:methods-distributional}) yields the lowest median EMD ($0.24$), while the single-score variant is within 0.01 of the best median MAE. Both variants leading on distributional fidelity confirms that the quantile-matching mechanism drives performance, not the output representation.

Other distributional methods fall short due to specific structural limitations:
\begin{itemize}
  \item \textbf{Cross-Entropy Matching:} Although it targets score distributions and improves EMD over full-batch baselines (Section~\ref{subsec:rq1-does-calibration-help}), it trails SQM significantly (median MAE $0.67$ vs.\ $0.53$; Wasserstein $0.32$ vs.\ $0.26$). Its KL objective is distance-blind. It penalizes misplaced probability mass regardless of bin distance and its affine restriction prevents true reshaping.
  \item \textbf{Mini-Batch Sliced Wasserstein OT:} While solving 1-D optimal transport, it operates only within isolated, fixed-size mini-batches rather than maintaining a continuously refined global estimate, leading to middling MAE and no early-stopping benefit.
\end{itemize}

Comparing the baselines reveals two axes governing performance: ordinal awareness and distributional alignment. Pointwise accuracy requires ordinal awareness (``ordinal-aware beats ordinal-blind''), as shown by isotonic regression beating cross-entropy matching on MAE and exact match despite ignoring score distributions entirely. Conversely, explicit distributional alignment is a direct route to distributional shape fidelity, as the three lowest Wasserstein distances all belong to distributional methods. SQM's advantage holds at a matched data budget: its streaming trajectory at the 100\% checkpoint, having seen the same data both baselines are fit on in full-batch mode, still leads both (Figure~\ref{fig:rq1-full-limited}). This rules out data budget as a primary confounder. We note this functions as a quasi-ablation at the property level rather than model internals, disentangling the impact of loss function objectives (e.g., $L_1$/EMD vs.\ isotonic regression's $L_2$ loss) and modeling capacity (e.g., nonparametric flexibility vs.\ cross-entropy matching's affine restriction).



\section{Limitations, Future Work, and Conclusions}
\label{sec:discussion}

\emph{Data scale and language.} All datasets use a 5-point Likert scale; human alignment can vary on coarser or finer scales \citep{li2026gradingscale}. HelpSteer2's native 0--4 scale is mapped to 1--5 via a standard offset. All prompts and data are English-only due to resource constraints, though the core calibration benefit should be language-agnostic.

\emph{Judge models.} We hold every model's weights frozen and vary
only the post-hoc calibration function (Section~\ref{sec:problem-setup}); how the judge was
trained, prompted, or fine-tuned, all known to affect judge-human alignment
\citep{zhu2023judgelm, kim2023prometheus, wang2024pandalm, golchin2025influences}, is out of
scope. Whether a better trained or prompted judge needs less calibration data is left to
future work. Two frontier-scale (GPT-5, Kimi-K3) and two smaller judges
(GPT-oss-20b, Qwen3-30b) are enough to check that the frontier-vs-small gap and calibration's effect
on it (Section~\ref{subsec:rq2-model-gap}) are not an artifact of one model, but too few to
characterize which model families are more calibration-friendly.
This paper set out to answer, under one shared experimental protocol, how much human-labeled data a
frozen LLM judge needs to be calibrated well, whether calibration lets smaller judges compete with
larger ones, and which calibration method to actually pick. We evaluated fifteen calibration
methods, spanning streaming, batch, and full-batch fitting regimes, across four judge models
and five datasets, and three findings stand out as the paper's main contributions.

\section{Conclusion}
\label{sec:conclusion}

First, a method fit incrementally does not need to wait for a large, fully collected anchor set to
be useful. Across methods and datasets, the typical point at which an incrementally-fit method
matches full-batch-level held-out MAE falls at roughly a quarter to a third of the labeled anchor
pool (23--34\%; Section~\ref{subsec:rq1-incremental-vs-full}), and a full-batch fit itself gains
almost nothing from being given ten times more data (a 4.3\% relative MAE improvement from 10\% to
100\% of the pool). Streaming quantile matching is the clearest individual example of this:
its own pooled trajectory beats every one of these data-limited full-batch checkpoints, including
the checkpoint fit on 100\% of the data, using far less labeled data to get there
(Section~\ref{subsec:rq1-incremental-vs-full}). Whether that data-efficiency advantage is shared by
every incremental method or is specific to a handful of them is not something we generalize across
the whole streaming and batch fitting modes; we report each method on its own merits rather than
as a group in Section~\ref{sec:results}, and mini-batch sliced Wasserstein OT, the other incremental
distribution-matching method we test, does not show the same early advantage.

Second, across all the axes we varied -- data budget, judge model, calibration method, and fitting
mode -- the single factor that most determines whether calibration succeeds is not any of these,
but how uniform the target task is. The five datasets we calibrate against are not interchangeable:
BiGGen Bench's per-instance, multi-capability rubric is uniformly harder to calibrate than the
other four datasets' single, fixed rubric per attribute, and no calibration method, data budget, or
judge model closes that gap. Task uniformity, in other
words, is a property of the dataset that calibration cannot fit its way around. Where a task
\emph{is} sufficiently uniform, however, the amount of good calibration data matters more than
which model produced the raw scores: given enough well-chosen anchors, a small, cheap judge can be
substituted for a much larger one. This same calibration budget closes 58--85\% of the raw MAE gap
between a small judge (gpt-oss-20b, qwen3-30b) and a large one (gpt-5, kimi-k3), roughly double the
32--43\% reduction calibration buys within a size tier (Section~\ref{subsec:rq2-model-gap}) --
though, as Section~\ref{subsec:rq2-model-gap} details, this gap-closing effect is not uniform
across methods: it holds in direction for every substantive method we test, but its magnitude
ranges from 27\% to 85\% depending on which method is used, and a poor method choice
(median scaling) can erase it almost entirely. A
well-calibrated small model does not just narrow the gap to a frontier judge -- with the right
method and enough anchors it can land within a few hundredths of MAE of the strongest large-model
judge, at a fraction of the inference cost. This reinforces two practical points: correct,
task-appropriate human annotation remains essential, since calibration corrects a judge's
relationship to its labels rather than manufacturing agreement where the rubric itself is
heterogeneous; and task separation -- fitting an independent calibration function per attribute or
rubric rather than one global correction across heterogeneous tasks -- is worth doing, because it
is exactly what keeps calibration from having to average away real differences in the
judge-vs-human relationship across tasks. Critically, this separation is not a deployment burden.
Because calibration is a lightweight, post-hoc, per-task mapping fit after the judge's weights are
already frozen, adding a new task or attribute, or swapping in a new judge model, only requires
re-fitting cheap calibration functions on a small anchor set, not retraining or re-provisioning
anything about the judge itself.

Third, this paper is, to our knowledge, the first to apply streaming quantile matching to
LLM-judge calibration, and it is the method that wins outright most often and by the widest
margin. The method itself is not new -- it is the classical P\textsuperscript{2} algorithm for
online quantile estimation \citep{jain1985p2} -- but its use for calibrating LLM-judge scores
against human ratings is. Pooled across datasets, models, and prompts, it is the best-performing
method on two of the three metrics we report and within 0.02 of the best on the third
(Section~\ref{subsec:rq1-does-calibration-help}), and it finishes first on eleven of the fifteen
dataset$\times$metric combinations we test, more than any other method
(Section~\ref{subsec:rq1-does-calibration-help}). It is also the single most reliable method by
that same record: it never finishes last on any of the fifteen combinations, a distinction it
shares only with Platt scaling and neural ODE calibration, both of which win far less often
(Section~\ref{subsec:rq3-best-stable}). More broadly, we
also compared, for the first time under one shared protocol, several well-established calibration
and distribution-matching algorithms from the broader machine learning literature that had not
previously been applied to LLM-judge calibration -- including Dawid--Skene noisy-rater aggregation,
conformal prediction, and mini-batch sliced Wasserstein optimal transport -- alongside methods
proposed specifically for this setting (Section~\ref{sec:related-work}). Dawid--Skene EM in
particular lands 6th of fifteen methods by overall rank (Table~\ref{tab:rq1-heldout-medians}),
ahead of several purpose-built LLM-judge calibration methods including streaming Bayesian
posterior update and neural ODE calibration, evidence that general-purpose calibration machinery
not originally designed with LLM judges in mind has more to offer this setting than it has so far
been credited with.

Taken together, these results give practitioners concrete, actionable guidance: which fitting mode
to use under a given data budget, how much labeled data is actually worth collecting, when a
smaller judge model is a safe substitute for a larger one, and which method to default to when the
target task is not known in advance (Section~\ref{sec:discussion}). They also point to open
research questions this paper does not resolve: devising calibration methods that jointly fit
across heterogeneous tasks or rubrics rather than requiring separate per-task functions, and
understanding more precisely what makes a task's rubric structure calibration-friendly in the first
place, remain open problems for future work on LLM-judge calibration.

\bibliographystyle{unsrtnat}  
\bibliography{references}  

\appendix

\section{Formal Definitions of Calibration Methods}
\label{app:formal-definitions}

Section~\ref{sec:methods} introduces calibration methods  and already gives
formal definitions of methods directly
(Section~\ref{subsec:methods-distributional}). Of the remaining, judge
evaluation \citep{landesberg2025cje}, SCOPE \citep{badshah2026scope}, and Neural ODE calibration
\citep{morandi2026debias} are cited directly against a paper that already formalizes an
LLM-judge calibration method, so we do not re-derive them here. For the remaining methods,
whose citations (where they exist) are to classifier calibration or image-retrieval techniques
from outside the LLM-judge setting, we give the exact correction $g_\phi$ each one fits.

\paragraph{Median scaling.} With running median $\bar{j}$ over anchors seen so far and
$u_i = j_i - \bar{j}$, $v_i = h_i - \bar{j}$,
\[
T = \frac{\sum_i u_i^2}{\sum_i u_i v_i}, \qquad g_\phi(j) = \bar{j} + \frac{j - \bar{j}}{T}.
\]

\paragraph{Few-shot exemplar calibration.} Given the current exemplar pool
$\mathcal{A} = \{(j_i, h_i)\}$ and a new score $j$, let $N_k(j) \subseteq \mathcal{A}$ be the
$k$ exemplars with smallest $|j - j_i|$; with weights $w_i = 1 / (|j - j_i| + \epsilon)$,
\[
g_\phi(j) = \frac{\sum_{i \in N_k(j)} w_i h_i}{\sum_{i \in N_k(j)} w_i}.
\]

\paragraph{Streaming linear regression.} An affine map fit by ordinary least squares,
\[
g_\phi(j) = \alpha + \beta j, \qquad
\beta = \frac{\sum_i (j_i - \bar{j})(h_i - \bar{h})}{\sum_i (j_i - \bar{j})^2}, \qquad
\alpha = \bar{h} - \beta \bar{j},
\]
with $(\alpha, \beta)$ recomputed after every anchor from running first/second moments and the
cross-moment (recursive least squares), not refit from scratch.

\paragraph{Streaming Bayesian posterior update.} With feature map
$\phi(j) = (j^2,\, j,\, 1)^\top$, weight vector $w$, prior precision $\Lambda_0 = \tau I$, and
observation noise variance $\sigma^2$, each new anchor $(j_i, h_i)$ updates the posterior
precision and precision-weighted mean in closed form,
\[
\Lambda_i = \Lambda_{i-1} + \frac{\phi(j_i)\phi(j_i)^\top}{\sigma^2}, \qquad
\Lambda_i \mu_i = \Lambda_{i-1}\mu_{i-1} + \frac{\phi(j_i)\, h_i}{\sigma^2},
\]
and $g_\phi(j) = \phi(j)^\top \mu_n$ after $n$ anchors.

\paragraph{Dawid--Skene EM.} Scores are discretized into $K$ classes. Let $\pi_k$ be the prior
over the latent true class $k$, and $\pi^J_{k,c}$, $\pi^H_{k,c}$ the judge's and (during fitting)
the human's confusion matrices, $P(\text{observed class}=c \mid \text{true}=k)$. EM alternates
\[
\gamma_i(k) \propto \pi_k\, \pi^J_{k, c_i^J}\, \pi^H_{k, c_i^H}
\quad\text{(E-step)}, \qquad
\pi_k \propto \sum_i \gamma_i(k), \quad
\pi^J_{k,c} \propto \sum_i \gamma_i(k)\, \mathbb{1}[c_i^J = c]
\quad\text{(M-step)},
\]
with the human update analogous. At prediction time only the judge's class $c(j)$ is observed, and
\[
g_\phi(j) = \sum_k \gamma(k \mid j)\, v_k, \qquad
\gamma(k \mid j) \propto \pi_k\, \pi^J_{k, c(j)},
\]
where $v_k$ is class $k$'s score midpoint.

\paragraph{Contrastive Rollout Calibration.}
Each judge output is a rollout over $K$ rating values $r_1,\ldots,r_K$, snapped onto the fixed
rating enum as a probability vector $p \in \Delta^{K-1}$ and mapped to logit space
$z = \log(p + \epsilon)$. During fitting the $(j,h)$ pairing is used directly: example $i$ is
\emph{agreeing} if $|\sum_k r_k p_{ik} - h_i| \le \tau$ and \emph{disagreeing} otherwise. Two
streaming means accumulate logits over the two groups,
\[
\mu_{\mathrm{pos}} = \mathbb{E}[z \mid \text{agree}], \qquad
\mu_{\mathrm{neg}} = \mathbb{E}[z \mid \text{disagree}], \qquad
v = \mu_{\mathrm{pos}} - \mu_{\mathrm{neg}}.
\]
At prediction time only the judge rollout is observed. Let $\hat{v} = v/\|v\|$,
$\mathrm{target} = \mu_{\mathrm{pos}} \cdot \hat{v}$, and $s = z \cdot \hat{v}$. The calibrated
logits are
\[
z' = z + \max(0,\ \mathrm{target} - s)\,\hat{v},
\]
leaving examples that already project at or beyond the agreeing cluster unchanged. The calibrated
score is
\[
g_\phi(j) = \sum_k r_k\,\mathrm{softmax}(z')_k,
\]
a convex combination of the rating enum and therefore always in $[\min_k r_k, \max_k r_k]$.

\paragraph{Conformal prediction.} A streaming-linear-regression point estimate
$\hat{h}(j) = \alpha + \beta j$ is maintained as above; over a sliding window of the $W$ most
recent nonconformity scores $r_i = |\hat{h}(j_i) - h_i|$, the margin is their
$(1-\alpha)$-quantile,
\[
g_\phi(j) = \hat{h}(j), \qquad
\text{margin} = \operatorname{Quantile}_{1-\alpha}\!\big(\{r_i\}_{i \in \text{window}}\big),
\]
with the returned interval $[\hat{h}(j) - \text{margin},\, \hat{h}(j) + \text{margin}]$ used only
for coverage evaluation, not for the point-estimate error metrics.

\paragraph{Platt scaling.} Human scores are min-max normalized to soft targets
$t_i = (h_i - h_{\min}) / (h_{\max} - h_{\min}) \in [\epsilon, 1-\epsilon]$, and $(A, B)$ are fit
by minimizing cross-entropy against $t_i$,
\[
(A, B) = \argmin_{A,B} -\sum_i \Big[t_i \log \sigma(Aj_i + B) + (1-t_i)\log\big(1-\sigma(Aj_i+B)\big)\Big],
\]
with $\sigma$ the logistic sigmoid, and
$g_\phi(j) = h_{\min} + (h_{\max} - h_{\min})\, \sigma(Aj + B)$.

\paragraph{Isotonic regression.} The non-parametric monotone least-squares fit,
\[
g_\phi = \argmin_{g \text{ monotone non-decreasing}} \sum_i \big(h_i - g(j_i)\big)^2,
\]
solved via the pool-adjacent-violators algorithm.

\paragraph{Score-debiased KDE (SD-KDE).} At query $j$, with Gaussian kernel weights
$w_i(j) = K\!\big((j - j_i)/h\big)$, a weighted local-linear fit
$(\beta_0(j), \beta_1(j)) = \argmin_{\beta_0,\beta_1} \sum_i w_i(j) \big(h_i - \beta_0 - \beta_1(j_i - j)\big)^2$
is solved at every query, and
\[
g_\phi(j) = \beta_0(j),
\]
the fitted local intercept -- this removes the leading-order boundary/curvature bias of a plain
Nadaraya--Watson weighted average $\sum_i w_i(j) h_i / \sum_i w_i(j)$.

\section{Supporting Material: Held-Out Metrics per Calibration Method}
\label{app:rq1-heldout-metrics}

Table~\ref{tab:rq1-heldout-medians} in Section~\ref{subsec:rq1-does-calibration-help} reports the
per-method, per-track results and the average, pooled effect of calibration across all methods. This section gives the full distributional view behind that
table.

\begin{figure}[t]
\centering
\includegraphics[width=\textwidth]{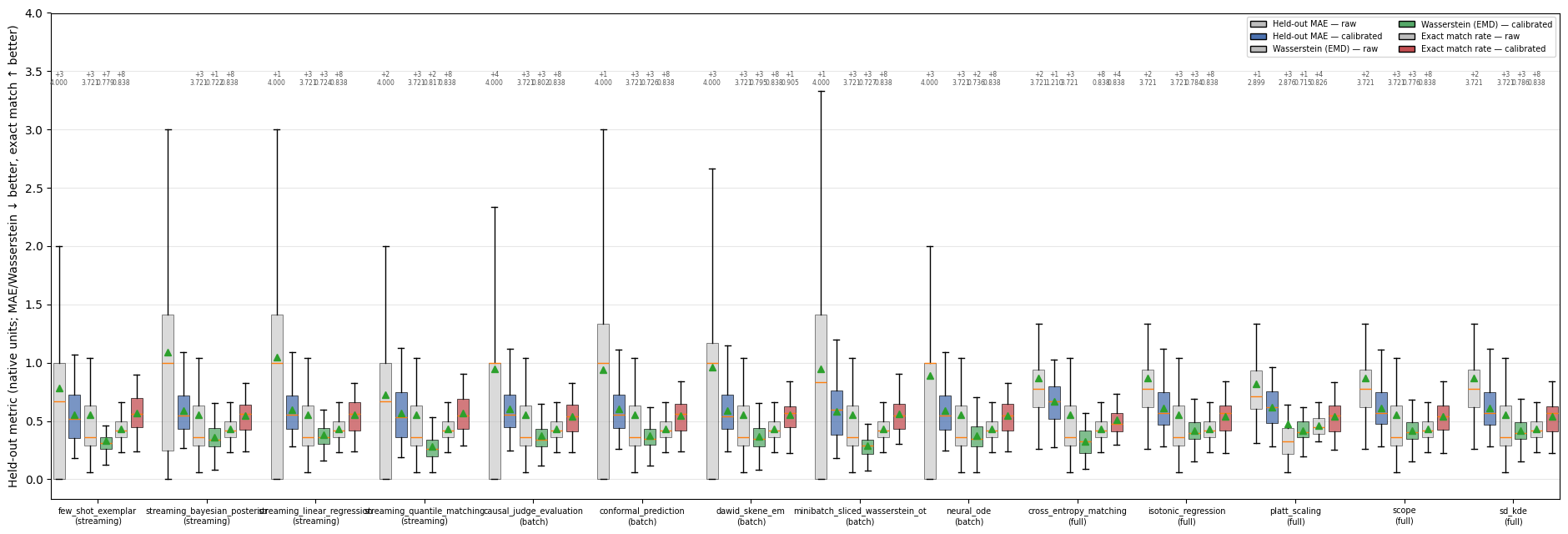}
\caption{Held-out MAE, Wasserstein distance, and exact-match rate per calibration method (raw vs.\
best-checkpoint calibrated), in native units. Methods are grouped by track (streaming, batch,
full-batch); outlier counts and the most extreme outlier value are annotated above each box.}
\label{fig:rq1-appendix-native}
\end{figure}

Figures~\ref{fig:rq1-appendix-native} and~\ref{fig:rq1-appendix-normalized} show the same held-out
MAE, Wasserstein distance, and exact-match rate, broken out per method (grouped by mode:
streaming, batch, full-batch) rather than pooled. For each method, the raw-judge and calibrated
distributions are shown as boxplots across all (dataset, model, attribute) held-out groups.
Figure~\ref{fig:rq1-appendix-native} reports each metric in its native units; because MAE,
Wasserstein distance, and exact-match rate live on different scales, comparing methods across
metrics is easier in Figure~\ref{fig:rq1-appendix-normalized}, where each metric is min--max
normalized to $[0,1]$ and exact-match rate is flipped to a mismatch rate ($1 - \text{exact match}$),
so that lower is always closer to human on every metric.


\begin{table}[t]
  \centering
  \caption{Methods' MAE, Wasserstein distance, and exact-match rate, pooled across
  all datasets, models, and prompts. \emph{Median} is the median calibrated value across held-out groups; \emph{Best} is that method's single best-performing held-out group. \textbf{Bold} marks the best value in each column and \underline{underline} the second best; ($\Delta$\,\%) is the relative improvement over the raw scores; \emph{W/L} is the number of dataset$\times$metric combinations in which the method finished first / last by pairwise majority vote over metrics, with ties counted.}
  \label{tab:rq1-heldout-medians}
  \resizebox{\textwidth}{!}{%
  \begin{tabular}{llrlrlrlr}
  \toprule
  & & \multicolumn{2}{c}{Held-out MAE $\downarrow$} & \multicolumn{2}{c}{Wasserstein $\downarrow$} & \multicolumn{2}{c}{Exact match $\uparrow$} & \\
  \cmidrule(lr){3-4} \cmidrule(lr){5-6} \cmidrule(lr){7-8}
  Method (Reference) & Mode & Median ($\Delta$ +/-\%) & Best ($\Delta$+/- \%) & Median ($\Delta$+/- \%) & Best ($\Delta$+/- \%) & Median ($\Delta$+/- \%) & Best ($\Delta$+/- \%) & Rank (W/L) \\
  \midrule
   \textbf{SQM methods} (Our Proposal) \\
   \midrule
 {Streaming quantile matching} &{Streaming} &{\underline{0.53 (+31.37)}} &{\underline{0.19 (+75.24)}} &{\underline{0.26 (+27.87)}} &{\textbf{0.06 (+83.87)}} &{\text{0.55 (+30.86)}} &{\textbf{0.91 (+117.27)}} &{\textbf{1 (7W/0L)}} \\
 {SQM over predictive score distributions} &{Streaming} &{\text{0.53 (+30.81)}} &{\text{0.20 (+74.10)}} &{\textbf{0.24 (+33.80)}} &{\textbf{0.06 (+82.75)}} &{\text{0.55 (+30.73)}} &{\underline{0.90 (+115.64)}} &{6 (2W/0L)} \\
 \midrule
 \midrule
 \textbf{CDF methods}\\

 \midrule
{Mini-batch sliced Wasserstein OT \citep{bonneel2015sliced}} &{Batch} &{\text{0.60 (+22.45)}} &{\textbf{0.18 (+76.29)}} &{\text{0.29 (+20.52)}} &{\underline{0.07 (+79.85)}} &{\text{0.55 (+30.82)}} &{\textbf{0.91 (+117.27)}} &{5 (5W/0L)} \\
 {Cross-entropy matching } &{Full-batch} &{\text{0.67 (+13.36)}} &{\text{0.28 (+64.11)}} &{\text{0.32 (+11.47)}} &{\text{0.09 (+75.58)}} &{\text{0.47 (+13.49)}} &{\text{0.84 (+100.50)}} &{17 (0W/3L)} \\
  Contrastive rollout calibration inspired by~\citep{rimsky2023caa} & Streaming & \text{0.65 (+16.07)} & \text{0.23 (+69.71)} & \text{0.34 (+7.71)} & \text{0.12 (+65.71)} & \text{0.48 (+15.48)} & \text{0.89 (+112.74)} & 15 (0W/1L) \\
  
 \midrule
 \midrule
 \textbf{Other methods} \\
 \midrule
  Dawid--Skene EM \citep{dawidskene1979} & Batch & \text{0.54 (+29.79)} & \text{0.24 (+68.87)} & \text{0.35 (+4.00)} & \text{0.08 (+77.65)} & \textbf{0.57 (+35.41)} & \textbf{0.91 (+116.68)} & \underline{2 (1W/0L)} \\
  Few-shot exemplar (inspired by \citep{zhao2021calibrate}) & Streaming & \textbf{0.52 (+32.86)} & \textbf{0.18 (+76.47)} & \text{0.31 (+15.27)} & \text{0.09 (+76.28)} & \underline{0.56 (+35.16)} & \underline{0.90 (+115.52)} & 3 (5W/1L) \\
  Streaming Bayesian posterior update \citep{bishop2006prml} & Streaming & \text{0.55 (+28.87)} & \text{0.27 (+65.06)} & \text{0.34 (+5.97)} & \text{0.08 (+77.65)} & \underline{0.56 (+33.80)} & \text{0.83 (+98.16)} & 4 (1W/0L) \\
 Causal judge evaluation \citep{landesberg2025cje} & Batch & \text{0.55 (+28.25)} & \text{0.24 (+68.39)} & \text{0.34 (+6.17)} & \text{0.12 (+67.35)} & \text{0.54 (+29.94)} & \text{0.83 (+97.51)} & 7 (1W/0L) \\
  Neural ODE calibration (inspired by) \citep{morandi2026debias} & Batch & \text{0.55 (+29.35)} & \text{0.24 (+68.34)} & \text{0.35 (+4.58)} & \textbf{0.06 (+82.37)} & \text{0.54 (+30.38)} & \text{0.83 (+97.51)} & 8 (1W/0L) \\
  Streaming linear regression \citep{young1984recursive} & Streaming & \text{0.55 (+28.47)} & \text{0.28 (+63.62)} & \text{0.36 (+0.82)} & \text{0.16 (+55.34)} & \underline{0.56 (+34.96)} & \text{0.83 (+97.51)} & 9 (2W/0L) \\
  Conformal prediction (derived from \citep{vovk2005conformal}) & Batch & \text{0.56 (+28.19)} & \text{0.26 (+66.67)} & \text{0.34 (+5.69)} & \text{0.11 (+68.55)} & \underline{0.56 (+33.66)} & \text{0.84 (+100.50)} & 10 (2W/0L) \\
  Median scaling & Streaming & \text{0.71 (+8.43)} & \text{0.26 (+66.67)} & 0.43 (-18.19) & \text{0.08 (+77.65)} & \text{0.45 (+8.54)} & \text{0.84 (+100.50)} & 11 (1W/3L) \\
  Isotonic regression \citep{zadrozny2002isotonic} & Full-batch & \text{0.57 (+26.17)} & \text{0.28 (+63.59)} & 0.40 (-8.61) & \text{0.15 (+58.72)} & \textbf{0.57 (+35.41)} & \text{0.84 (+100.50)} & 12 (0W/1L) \\
  SCOPE\citep{badshah2026scope} & Full-batch & \text{0.57 (+26.38)} & \text{0.28 (+63.46)} & 0.40 (-10.13) & \text{0.16 (+57.31)} & \text{0.54 (+30.02)} & \text{0.84 (+100.50)} & 13 (0W/0L) \\
  Score-debiased KDE (SD-KDE) \citep{epstein2025sdkde} & Full-batch & \text{0.57 (+26.19)} & \text{0.28 (+63.59)} & 0.39 (-8.23) & \text{0.15 (+58.72)} & \textbf{0.57 (+35.41)} & \text{0.84 (+100.50)} & 14 (0W/1L) \\
 Platt scaling \citep{platt1999} & Full-batch & \text{0.62 (+20.08)} & \text{0.29 (+63.12)} & 0.41 (-11.27) & \text{0.20 (+46.32)} & \text{0.55 (+30.83)} & \text{0.83 (+99.21)} & 16 (2W/5L) \\
 
  \bottomrule
  \end{tabular}%
  }
  \end{table}

\section{Supporting Material: Per-Dataset and Per-Model Method Comparison}
\label{app:rq3-method-ranking}

\begin{figure}[t]
\centering
\includegraphics[width=\textwidth]{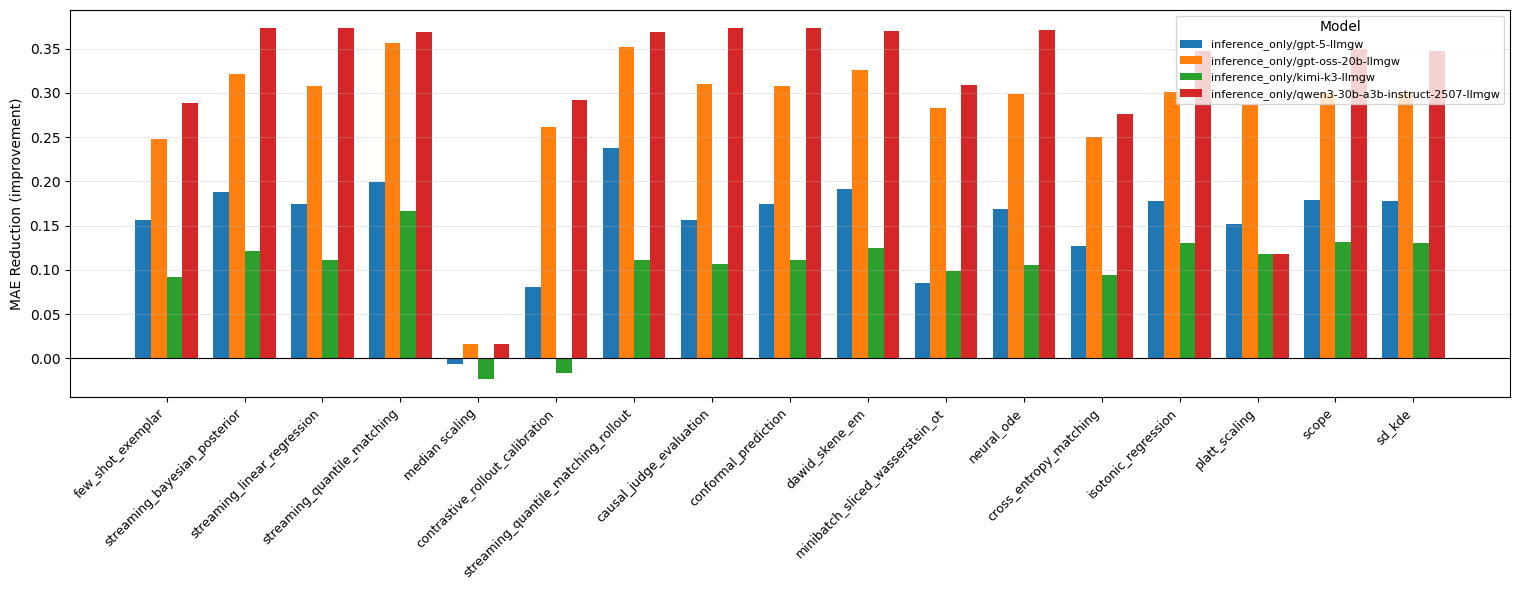}
\caption{Held-out MAE reduction (raw judge minus calibrated final MAE) per calibration method, one
bar cluster per judge model. Larger is better. Improvement is consistently largest for the two
weaker judges (gpt-oss-20b, qwen3-30b) and smallest for kimi-k3; median scaling is the
exception, with near-zero improvement for every model.}
\label{fig:rq3-improvement-by-model}
\end{figure}

Section~\ref{subsec:rq3-best-stable} looks at each method's held-out MAE reduction split out
by judge model, and at its stability using the win/last-place record from
Table~\ref{tab:rq1-heldout-medians}. Figure~\ref{fig:rq3-improvement-by-model} gives the per-model
breakdown

\end{document}